\documentclass[10pt,letterpaper]{article}

\usepackage[margin=1in]{geometry}

\usepackage[utf8]{inputenc}
\usepackage[T1]{fontenc}
\usepackage{lmodern}
\usepackage{hyperref}
\usepackage{url}
\usepackage{booktabs}
\usepackage{amsfonts}
\usepackage{amsmath, amssymb}
\usepackage{amsthm}
\usepackage{mathtools}
\usepackage{nicefrac}
\usepackage{microtype}
\usepackage{fancyhdr}
\usepackage{graphicx}
\graphicspath{{figures/}}
\usepackage{enumerate}
\usepackage{enumitem}
\usepackage{bm}
\usepackage{bbm}
\usepackage{natbib}

\newcommand{\keywords}[1]{\par\smallskip\noindent\textbf{Keywords:} #1\par}

\newtheorem{proposition}{Proposition}
\newtheorem{lemma}{Lemma}

\newtheorem{definition}{Definition}

\title{Equilibrium Selection in Multi-Agent Policy Gradients\\
via Opponent-Aware Basin Entry}

\author{%
  Yevhen Shcherbinin$^{1,2}$ \quad
  Arina Redina$^{3,1}$ \quad
  Maxim Kalpin$^{1,4}$ \quad
  Vlad Kochetov$^{1,5}$
  \\[0.5em]
  \small $^1$Bloomsbury Technology, London, United Kingdom\\
  \small $^2$London School of Economics and Political Science, London, United Kingdom\\
  \small $^3$University of Bristol, Bristol, United Kingdom\\
  \small $^4$Johannes Kepler University Linz, Linz, Austria\\
  \small $^5$Odesa Polytechnic National University, Odesa, Ukraine\\[0.3em]
  \small Correspondence: \texttt{eugene.shcherbinin@bloomsburytech.com}
}

\date{}

\begin{document}
\maketitle
\thispagestyle{empty}

\begin{abstract}
Multi-agent policy-gradient methods have been shown to converge locally near stable Nash equilibria. Local convergence, however, does not determine which equilibrium is reached. We study this question through basin-entry probability with respect to a target set of equilibria selected by an external criterion, such as payoff dominance. For finite-unroll Meta-MAPG, we show that the update decomposes into ordinary policy gradient plus own-learning and peer-learning corrections, with controlled sampling noise and finite-unroll bias. We identify the peer-learning correction as the main equilibrium-selection mechanism: under a local alignment condition, the probability of entering the certified attraction region of the target stable-Nash set increases, relative to ordinary policy gradient. Because persistent correction may shift zero-update points of the original game, annealing the correction after entering the basin recovers ordinary policy-gradient dynamics and inherits local stable-Nash convergence guarantees. Experiments in Stag Hunt, iterated Prisoner's Dilemma, and preliminary neural-policy coordination environments support this basin-entry view, showing increased entry into cooperative basins under peer-aware updates.
\end{abstract}


\section{Introduction}

\citet{giannou2022convergence} established that multi-agent policy gradient converges locally near stable Nash equilibria: from initialisations sufficiently close to one, the dynamics enter and stay. However, generic stochastic games admit multiple stable Nash equilibria with very different welfare, coordination, and safety properties, and local convergence is silent on which one learning reaches from an arbitrary initialisation. It is then natural to ask under what conditions the probability of entering the attraction region of a preferred equilibrium can be increased. We reframe this as a \emph{finite-time basin-entry problem}: an external criterion $W$ --- payoff dominance, welfare, cooperation --- designates a target set of stable Nash equilibria $\mathcal{N}^\star_W$, and we ask whether a learning algorithm can raise the probability that iterates enter their certified attraction region within a finite horizon.

Steering toward a chosen basin requires reasoning about how a policy shapes the future learning of its peers. Meta-MAPG \citep{kim2021meta}, originally developed for rapid adaptation, differentiates through both an agent's own future updates and those of every other agent. We reinterpret it as a basin-entry mechanism and show that its finite-unroll update decomposes into ordinary policy gradient plus own- and peer-learning corrections, with the sampled estimator admitting a stochastic-approximation decomposition with martingale-difference noise and summable finite-unroll bias. The own-learning correction vanishes at Nash while the peer-learning correction need not, making the peer term the load-bearing equilibrium-selection mechanism. Under a local alignment condition, this correction expands the certified attraction region, increasing target-basin entry probability relative to plain policy gradient.

Because peer shaping displaces the fixed point from the Nash equilibria of the original game, we introduce a \emph{shape-then-cool} schedule that anneals the correction after basin entry to recover ordinary policy-gradient dynamics and the local convergence guarantees of \citet{giannou2022convergence}. In tabular Stag Hunt, Meta-MAPG expands the cooperative basin from $27.0\%$ under PG to $42.6\%$, with ablations confirming the gain is carried by the peer term. The same dynamics holds in IPD where peer-aware methods reach $32$--$37\%$ cooperative success against $8\%$ for PG and own-only. Preliminary neural-policy experiments are noisier and tail-driven, but broadly support that peer-aware shaping alters equilibrium selection near basin boundaries.

\section{Related Work}
\label{sec:related}

\paragraph{Policy-gradient convergence in stochastic games.}
Recent work establishes almost-sure local convergence of policy gradient to stable Nash equilibria \citep{giannou2022convergence}, but these guarantees are conditional on attraction-region entry. We study the complementary finite-time question: whether opponent-aware updates can increase entry into a criterion-selected target basin.

\paragraph{Opponent-aware learning.}
Opponent-aware learning differentiates through anticipated opponent updates, beginning with LOLA \citep{foerster2018lola} and followed by stable, consistent, and proximal refinements such as SOS, COLA, and POLA \citep{letcher2019stable,willi2022cola,zhao2022pola}. Meta-MAPG \citep{kim2021meta} extends this family to finite multi-step unrolls in multi-agent reinforcement learning. We use this line of work as the algorithmic basis for studying how own- and peer-learning corrections affect basin-entry geometry.

\paragraph{Mechanism-facing MARL experiments.}
Most evaluations of opponent-aware methods report final returns on neural benchmarks, leaving basin identity inferential. We therefore use classic tabular coordination games, Stag Hunt and iterated Prisoner's Dilemma, where basin geometry can be inspected directly, and treat neural benchmarks such as Overcooked, MPE, and PettingZoo IPD as preliminary transfer probes rather than the main evidence.

\section{Problem Setup}
\label{sec:setup}

\subsection{Stochastic Games and Policies}

We consider $N$-player finite discounted stochastic games. A game is specified by a tuple:
\[
\mathcal{G} = \left(\mathcal{S},\, \mathcal{N},\, \{\mathcal{A}_i, R_i\}_{i \in \mathcal{N}},\, P,\, \gamma_{\mathrm{disc}},\, \rho_0\right),
\]
where $\mathcal{S} = \{1, \ldots, S\}$ is a finite state space,
$\mathcal{N} = \{1, \ldots, N\}$ is the set of agents,
$\mathcal{A}_i = \{1, \ldots, A_i\}$ is agent $i$'s action set with joint space
$\mathcal{A} = \prod_{i \in \mathcal{N}} \mathcal{A}_i$ and opponent space
$\mathcal{A}_{-i} = \prod_{j \neq i} \mathcal{A}_j$,
$R_i \colon \mathcal{S} \times \mathcal{A} \to [-1, 1]$ is agent $i$'s reward function,
$P(s' \mid s, a)$ is the transition kernel,
$\gamma_{\mathrm{disc}} \in [0, 1)$ is the discount factor so that agent $i$'s return is
$\sum_{t \geq 0} \gamma_{\mathrm{disc}}^{\,t}\, r_{i,t}$,
and $\rho_0 \in \Delta(\mathcal{S})$ is the initial state distribution.

Each agent $i$ plays a stationary Markovian policy $\pi_{\phi_i}(a_i \mid s)$, twice differentiable in its parameter $\phi_i \in \mathbb{R}^{d_i}$. The joint parameter $\phi = (\phi_i)_{i=1}^N \in \mathbb{R}^d$ is the single object all updates act on. Agent $i$'s value function $V_{i,\rho}(\phi)$ is the expected discounted return under joint policy $\phi$,
\begin{equation}
    V_{i,\rho}(\phi) := \mathbb{E}_{p(\tau \mid \phi)}\!\left[\,
    \sum_{t \geq 0} \gamma_{\mathrm{disc}}^{\,t}\, R_i(s_t, a_t)\right],
\end{equation}
which each agent seeks to maximise. Because agents optimise their own $V_{i,\rho}$ simultaneously, no agent controls the full environment: every policy update by agent $i$ changes the returns of all other agents. This interdependence is what makes equilibrium selection nontrivial.

The ordinary policy-gradient (PG) update for the joint parameter $\phi$ is
\begin{equation}
    \phi_{n+1} = \phi_n + \alpha_n\, v(\phi_n), \qquad
    v(\phi) := \bigl(\nabla_{\phi_i} V_{i,\rho}(\phi)\bigr)_{i=1}^N,
\end{equation}
where each component $\nabla_{\phi_i} V_{i,\rho}(\phi)$ is agent $i$'s individual policy gradient and $\alpha_n > 0$ is the step size.

\subsection{Nash Equilibria and Certified Attraction Regions}

The central solution concept in stochastic games is the Nash policy: a joint policy from which no agent can improve their return by unilaterally deviating. Formally,

\begin{definition}[Nash Policy]
A joint policy $\phi^* = (\phi^*_i)_{i=1}^N$ is a Nash policy if for every agent $i \in \mathcal{N}$ and all $\phi'_i \in \mathbb{R}^{d_i}$,
\begin{equation}
    V_{i,\rho}(\phi^*_i, \phi^*_{-i}) \geq V_{i,\rho}(\phi'_i, \phi^*_{-i}).
\end{equation}
\end{definition}

\citet{fink1964equilibrium} established that every finite discounted stochastic game admits at least one stationary Nash policy, so the solution concept is non-vacuous. However, existence alone does not guarantee that learning dynamics converge to a Nash policy. Some Nash policies are unstable under gradient updates, and learning algorithms may diverge away from them even when initialised nearby. We therefore work with stronger refinements that admit local stable-convergence guarantees.

\begin{definition}[Stable Nash and SOS \citep{giannou2022convergence}]
A Nash policy $\phi^*$ is stable if
\begin{equation}
    \langle v(\phi),\, \phi - \phi^* \rangle < 0
\end{equation}
for all $\phi \neq \phi^*$ sufficiently close to $\phi^*$, and second-order stationary (SOS) if there exist $\mu > 0$ and $\rho > 0$ such that
\begin{equation}
    \langle v(\phi),\, \phi - \phi^* \rangle \leq -\mu\|\phi - \phi^*\|^2
\end{equation}
whenever $\|\phi - \phi^*\| \leq \rho$.
\end{definition}

These conditions are nested --- SOS implies stable, and stable implies Nash. A finite stochastic game can admit multiple stable Nash equilibria, each with its own certified attraction region $B_{r_{\mathrm{att}}}(\phi^*)$ --- the set of initialisations from which \citet{giannou2022convergence} guarantee convergence of ordinary PG.

\begin{definition}[Certified attraction region]
Let $\phi^*$ be an SOS Nash equilibrium with constants $\mu > 0$ and $r_{\mathrm{att}} > 0$. The certified attraction region of $\phi^*$ is the open ball
\begin{equation}
    B_{r_{\mathrm{att}}}(\phi^*) := \{\phi \in \mathbb{R}^d : \|\phi - \phi^*\|
    < r_{\mathrm{att}}\}
\end{equation}
on which the SOS drift condition holds.
\end{definition}

\begin{lemma}[Local PG convergence \citep{giannou2022convergence}]
\label{lem:local}
If $\phi_0 \in B_{r_{\mathrm{att}}}(\phi^*)$ for an SOS Nash equilibrium $\phi^* \in \mathcal{E}_{\mathrm{stab}}$, and the step sizes satisfy $\sum_n \alpha_n = \infty$, $\sum_n \alpha_n^2 < \infty$, then $\phi_n \to \phi^*$ almost surely.
\end{lemma}

Local convergence theory is therefore conditional on being inside the right region to begin with, and does not determine which equilibrium learning reaches from an arbitrary initialisation. We formalise equilibrium selection by introducing an external criterion $W$ --- such as payoff dominance or social welfare --- that identifies a target subset of stable Nash equilibria,
\begin{equation}
    \mathcal{N}^{\star}_{W} := \{\phi^* \in \mathcal{E}_{\mathrm{stab}} : W(\phi^*) \geq \tau\},
\end{equation}
where $\mathcal{E}_{\mathrm{stab}}$ denotes the set of all stable Nash equilibria of the game. The target attraction region is the union of certified attraction regions of equilibria in $\mathcal{N}^{\star}_{W}$,
\begin{equation}
    \mathcal{B}_W := \bigcup_{\phi^* \in \mathcal{N}^{\star}_{W}} B_{r_{\mathrm{att}}}(\phi^*).
\end{equation}
The criterion $W$ defines which runs count as successes; it does not modify any update step.

The central object of this paper is the probability that a learning algorithm enters the certified attraction region of the target set $\mathcal{N}^{\star}_{W}$ within a fixed horizon $T \geq 1$.

\begin{definition}[Basin-entry probability]
For an algorithm $A$, horizon $T \geq 1$, and target set $\mathcal{N}^{\star}_{W}$, the basin-entry probability is
\begin{equation}
    p_{\mathrm{entry}}(A, T; W) := \Pr\bigl(\phi_T \in \mathcal{B}_W\bigr),
\end{equation}
where $\phi_T$ is the iterate produced by $A$ after $T$ steps from an initial policy $\phi_0 \in \mathbb{R}^d$.
\end{definition}

We note that the event $\phi_T \in \mathcal{B}_W$ is a finite-time event: it asks whether the iterate has reached a region where local convergence theory applies; what happens after entry is addressed in Section~\ref{sec:cooldown}. This is distinct from asymptotic convergence, and it is also distinct from the event of entering any stable-Nash basin, which depends on game geometry and the initialisation distribution alone. The central question is whether opponent-aware updates improve target basin entry relative to plain PG,
\begin{equation}
    p_{\mathrm{entry}}(\mathrm{Meta\text{-}MAPG},\, T)
    > p_{\mathrm{entry}}(\mathrm{PG},\, T).
\end{equation}

\section{The Meta-MAPG Update Decomposition}
\label{sec:decomposition}

Ordinary PG treats each agent's peers as a fixed background, ignoring simultaneous learning. \citet{kim2021meta} propose Meta-MAPG to model this interdependence explicitly, differentiating through both an agent's own future updates and the future updates of its peers. \citet{kim2021meta} develop Meta-MAPG as a rapid-adaptation algorithm; we reinterpret it as a candidate basin-entry mechanism.

\subsection{Finite-Unroll Meta-MAPG}

We model learning as a finite chain of $L$ inner-loop policy-gradient updates,
\begin{equation}
    \phi^i_{\ell+1} = \phi^i_\ell + \alpha^i \nabla_{\phi^i_\ell} V_{i,\rho}(\phi_\ell),
    \qquad
    \phi^{-i}_{\ell+1} = \phi^{-i}_\ell + \alpha^{-i}
    \nabla_{\phi^{-i}_\ell} V_{-i,\rho}(\phi_\ell),
\end{equation}
where $V_{-i,\rho}(\phi) := (V_{j,\rho}(\phi))_{j \neq i}$ is the tuple of value functions of all agents except $i$, and $\alpha^{-i}$ is their shared step size.

The meta-gradient of agent $i$ with respect to its initial parameters $\phi^i_0$ decomposes as
\begin{align}
  &\nabla_{\phi^i_0} V^i_{0:L}(s_0,\phi^i_0) \notag\\
  &\quad= \mathbb{E}_{p(\tau_{0:L-1}|\phi_{0:L-1})}\!\Bigl[
      \mathbb{E}_{p(\tau_{L}|\phi_{L})}\!\bigl[G^i(\tau_{L})\bigr]
      \Bigl(
      \underbrace{\nabla_{\phi^i_0}\log\pi(\tau_0|\phi^i_0)}_{\text{current}}
      \notag\\
  &\qquad\qquad
      +\underbrace{\sum_{\ell=0}^{L-1}
        \nabla_{\phi^i_0}\log\pi(\tau_{\ell+1}|\phi^i_{\ell+1})}_{\text{own-learning}}
      +\underbrace{\sum_{\ell=0}^{L-1}
        \nabla_{\phi^i_0}\log\pi(\tau_{\ell+1}|\phi^{-i}_{\ell+1})}_{\text{peer-learning}}
      \Bigr)\Bigr],
\end{align}
where $G^i(\tau_L) = \sum_{t \geq 0} \gamma_{\mathrm{disc}}^t R_i(s_t, a_t)$ is agent $i$'s discounted return along $\tau_L$.

We write the Meta-MAPG correction as
\begin{equation}
    M_L(\phi) := M^{\mathrm{own}}_L(\phi) + M^{\mathrm{peer}}_L(\phi),
\end{equation}
where $M^{\mathrm{own}}_L(\phi)$ collects the own-learning terms and $M^{\mathrm{peer}}_L(\phi)$ collects the peer-learning terms. The full Meta-MAPG update is
\begin{equation}
    F_L(\phi) := v(\phi) + M_L(\phi).
\end{equation}

The analysis focuses on finite unroll length $L < \infty$; infinite unrolls are computationally intractable, and finite truncation introduces a controllable bias whose summability under a growing schedule $L_n$ is established in the proof of Proposition~\ref{prop:sa}.

\subsubsection{Own-Learning vs.\ Peer-Learning}
\label{sec:ownvspeer}

The two correction terms in $M_L(\phi)$ play different conceptual roles. The own-learning term captures how agent $i$'s current policy shapes its own future adaptation. The peer-learning term captures how agent $i$'s current policy influences the future updates of its peers. At a Nash equilibrium $\phi^*$, each agent's policy gradient vanishes, so $\nabla_{\phi^i_0} V_{i,\rho}(\phi^*) = 0$ and the own-learning correction is zero at $\phi^*$. The peer-learning term, by contrast, does not generally vanish at Nash, and can therefore shift the certified attraction region of $\phi^*$ --- this is why peer-learning drives basin entry while own-learning does not. Peer-learning is therefore the natural candidate for the equilibrium-selection mechanism. This distinction is tested directly in our experiments in Section~\ref{sec:tabular-experiments}, which compare PG, own-learning-only, peer-learning-only, and full Meta-MAPG to isolate the peer term's contribution to basin entry.

\subsection{Stochastic-Approximation Decomposition}

Throughout this paper we work under the following regularity conditions.
\begin{itemize}[leftmargin=2em]
\item[\textbf{(RC1)}] Rewards are bounded: $|r_i| \leq R_{\max}$.
\item[\textbf{(RC2)}] Policies $\pi_{\phi_i}$ are twice differentiable in $\phi_i$ with score function $S_i(\tau) = \nabla_{\phi_i} \log p_\phi(\tau)$ and Hessian $H_i(\tau) = \nabla^2_{\phi_i} \log p_\phi(\tau)$ uniformly bounded on $K$.
\item[\textbf{(RC3)}] Likelihood-ratio estimators have finite second moments.
\item[\textbf{(RC4)}] Unroll length $L_n$ grows sufficiently fast that the truncation bias is summable.
\end{itemize}

The following proposition establishes that the sampled finite-unroll Meta-MAPG estimator has the noise and bias structure required to apply standard stochastic-approximation theory.

\begin{proposition}[Meta-MAPG as stochastic approximation]
\label{prop:sa}
The sampled finite-unroll Meta-MAPG update admits the decomposition
\begin{equation}
        g_n = v(\phi_n) + \lambda_n M_{L_n}(\phi_n) + \xi_{n+1} + b_n,
\end{equation}
where $M_{L_n} = M_{L_n}^{\mathrm{own}} + M_{L_n}^{\mathrm{peer}}$, $\{\xi_{n+1}\}$ is a martingale-difference sequence with $\mathbb{E}[\xi_{n+1} \mid \mathcal{F}_n] = 0$ and $\mathbb{E}[\|\xi_{n+1}\|^2 \mid \mathcal{F}_n] \leq \sigma^2$, and $\|b_n\| \leq C\beta_n$ for a deterministic sequence $\beta_n \to 0$.
\end{proposition}

The full proof is in Appendix~\ref{app:sa}.

\section{Opponent-Aware Basin Entry}
\label{sec:basin}

Local convergence theory, as established in Section~\ref{sec:setup}, does not characterise which basin learning enters from an arbitrary initialisation. When multiple stable Nash equilibria coexist, this leaves open which equilibrium is reached. In this section we characterise how the peer-learning correction shapes the certified attraction region geometry and increases the probability of entering the criterion-selected target basin $\mathcal{B}_W$.

\subsection{Local Alignment Condition}

The peer-learning correction improves basin entry when it is locally aligned with movement toward the target equilibrium. We formalise this through the Jacobian of the peer-learning correction evaluated at $\phi^*$. Define
\begin{equation}
    \mu_M := -\lambda_{\max}\!\left(\frac{J_M + J_M^\top}{2}\right),
    \qquad J_M := DM_L(\phi^*),
\end{equation}
where $\lambda_{\max}$ denotes the largest eigenvalue of the symmetrised Jacobian of $M_L$ at $\phi^*$. Under the local alignment condition $\mu_M > 0$, the certified attraction region strictly enlarges, as the following proposition makes precise.

\subsection{Local Geometry Under Peer-Learning Correction}

We state the result for a single target equilibrium $\phi^* \in \mathcal{N}^{\star}_W$; the same argument applies locally around each equilibrium in $\mathcal{N}^{\star}_W$. We characterise how adding the peer-learning correction $\lambda M_L$ to the base policy-gradient update changes the local geometry around $\phi^*$ in three concrete ways: it shifts the fixed point to a nearby point $\phi^*_\lambda$, improves the rate at which iterates are pulled toward it (drift improvement), and enlarges the certified attraction region when $\mu_M > 0$.

\begin{proposition}[Basin geometry under peer learning]
\label{prop:basin}
Fix an SOS Nash equilibrium $\phi^*$ of $v$ with drift constant $\mu > 0$, and suppose $v, M_L \in C^1$ near $\phi^*$. Let $F_\lambda(\phi) := v(\phi) + \lambda M_L(\phi)$ at shaping strength $\lambda \geq 0$, and let $L_\lambda$ denote the Lipschitz constant of $DF_\lambda$ on $B_{\rho_\lambda}(\phi^*_\lambda)$.
\begin{enumerate}[label=(\roman*)]
\item \textbf{Fixed-point shift.} $F_\lambda$ has its zero at a shifted point $\phi^*_\lambda$ satisfying $\|\phi^*_\lambda - \phi^*\| = O(\lambda)$.
\item \textbf{Drift improvement.} When $\mu_M > 0$, $F_\lambda$ contracts toward $\phi^*_\lambda$ with an improved rate:
\[
\langle F_\lambda(\phi),\, \phi - \phi^*_\lambda \rangle
\leq -\tfrac{1}{2}(\mu + \lambda\mu_M)\|\phi - \phi^*_\lambda\|^2.
\]
\item \textbf{Certified basin expansion.} The SOS ball around $\phi^*_\lambda$,
\[
B_{\rho_\lambda}(\phi^*_\lambda) := \bigl\{\phi \in \mathbb{R}^d :
\|\phi - \phi^*_\lambda\| < \rho_\lambda\bigr\},
\qquad \rho_\lambda := \frac{\mu + \lambda\mu_M}{2L_\lambda},
\]
has strictly larger certified radius than $B_{r_{\mathrm{att}}}(\phi^*)$ when $\mu_M > 0$ and $\lambda$ is sufficiently small, and certifies additional points near $\phi^*_\lambda$, enlarging the certified attraction region.
\end{enumerate}
\end{proposition}

The full proof is in Appendix~\ref{app:basin}.

By enlarging the certified attraction region of $\phi^*$, the peer-learning correction steers additional initialisations into $\mathcal{B}_W$. Provided the initialisation distribution charges the enlarged certified region with strictly greater probability than the original ball, this yields
\begin{equation}
    p_{\mathrm{entry}}(\mathrm{Meta\text{-}MAPG}, T)
    > p_{\mathrm{entry}}(\mathrm{PG}, T).
\end{equation}
Basin entry is therefore the right object to optimise for equilibrium selection, and the peer-learning correction is the mechanism that can improve it.

\section{Cooldown and Local Nash Convergence}
\label{sec:cooldown}

Proposition~\ref{prop:basin} certifies that the peer correction enlarges the target basin during warm-up, but does not address the asymptotic target after entry. The reason this matters is the same asymmetry that made the peer-learning term useful in the first place: because the peer correction does not vanish at Nash, holding it fixed permanently displaces the zero-update point. The stationary points of
\begin{equation}
    v(\phi) + \lambda M_L(\phi) = 0
\end{equation}
do not coincide with those of $v$, and Proposition~\ref{prop:basin}(i) quantifies the displacement as $\|\phi^*_\lambda - \phi^*\| = O(\lambda)$. Left on, the correction that steered the iterate into the right basin leads Meta-MAPG to a perturbed equilibrium rather than a Nash equilibrium of the original game.

A shape-then-cool schedule resolves this. During warm-up,
\begin{equation}
    \phi_{n+1} = \phi_n + \alpha_n\bigl(\hat{v}_n + \lambda_n
    \widehat{M}_{L_n,n}\bigr),
\end{equation}
with $\lambda_n$ non-negligible; after a handoff time $T$, $\lambda_n$ is either set to zero or annealed so that $\sum_{n \geq T} \alpha_n \lambda_n < \infty$. The handoff can be fixed in advance or driven by an observable statistic such as cooperation rate or task return, and does not require an oracle that identifies basin entry during training. Proposition~\ref{thm:cooldown} below formalises this: provided the iterate has entered $B_{r_{\mathrm{att}}}(\phi^*)$ by time $T$, annealing the correction recovers ordinary policy-gradient dynamics and inherits its local stable-Nash convergence guarantee.

\begin{proposition}[Local convergence after cooldown]
\label{thm:cooldown}
Assume the sampled update satisfies the decomposition in Proposition~\ref{prop:sa}. Let $\phi^*$ be an SOS Nash equilibrium of $v$ with certified attraction region $B_{r_{\mathrm{att}}}(\phi^*)$. Suppose $\phi_T \in B_{r_{\mathrm{att}}}(\phi^*)$ for some handoff time $T$ and the post-handoff iterates remain in the local certified region on which the SOS drift condition holds. Suppose further that $M_{L_n}$ is uniformly bounded on this region, $\|b_n\|\leq C\beta_n$ with $\sum_{n\geq T}\alpha_n\beta_n<\infty$, and the step sizes satisfy
\[
    \sum_{n \geq T} \alpha_n = \infty,
    \qquad
    \sum_{n \geq T} \alpha_n^2 < \infty.
\]
If either $\lambda_n = 0$ for all $n \geq T$, or $0\leq \lambda_n\leq \bar\lambda<\infty$ and
\[
    \sum_{n \geq T} \alpha_n \lambda_n < \infty,
\]
then $\phi_n \to \phi^*$ almost surely.
\end{proposition}

The full proof is in Appendix~\ref{app:cooldown}.

Together, Proposition~\ref{prop:basin} and Proposition~\ref{thm:cooldown} give a complete account of what Meta-MAPG converges to: the first characterises which basin the warm-up enters; the second guarantees that, once entry has occurred, the asymptotic target --- the original Nash equilibrium --- is recovered. The peer correction does its work during warm-up and is then annealed away, leaving ordinary policy-gradient dynamics to deliver the convergence guarantee. In tabular Stag Hunt, shape-then-cool reaches $42.5\%$ cooperative success against $41.25\%$ for constant Meta-MAPG, with second-half cooperation standard deviations below $5\times 10^{-4}$ for both arms. The empirical penalty for keeping the correction on is undetectable at this horizon --- but the theoretical guarantee requires the schedule.

\section{Tabular Experiments}
\label{sec:tabular-experiments}

Most of the objects used above, such as basins, separatrices, and the local geometry around a Nash equilibrium, are inferential. In Stag Hunt with Bernoulli policies they are visible. The policy space is the unit square, the two stable Nash equilibria sit at opposite corners, and the separatrix between their attraction regions is a curve we can draw. IPD gives the complementary check: its geometry is higher-dimensional, but ordinary policy gradient is close to the cooperation floor, so the peer-vs-non-peer comparison has a larger margin in a different game family.

\begin{figure}[t]
\centering
\includegraphics[width=\linewidth]{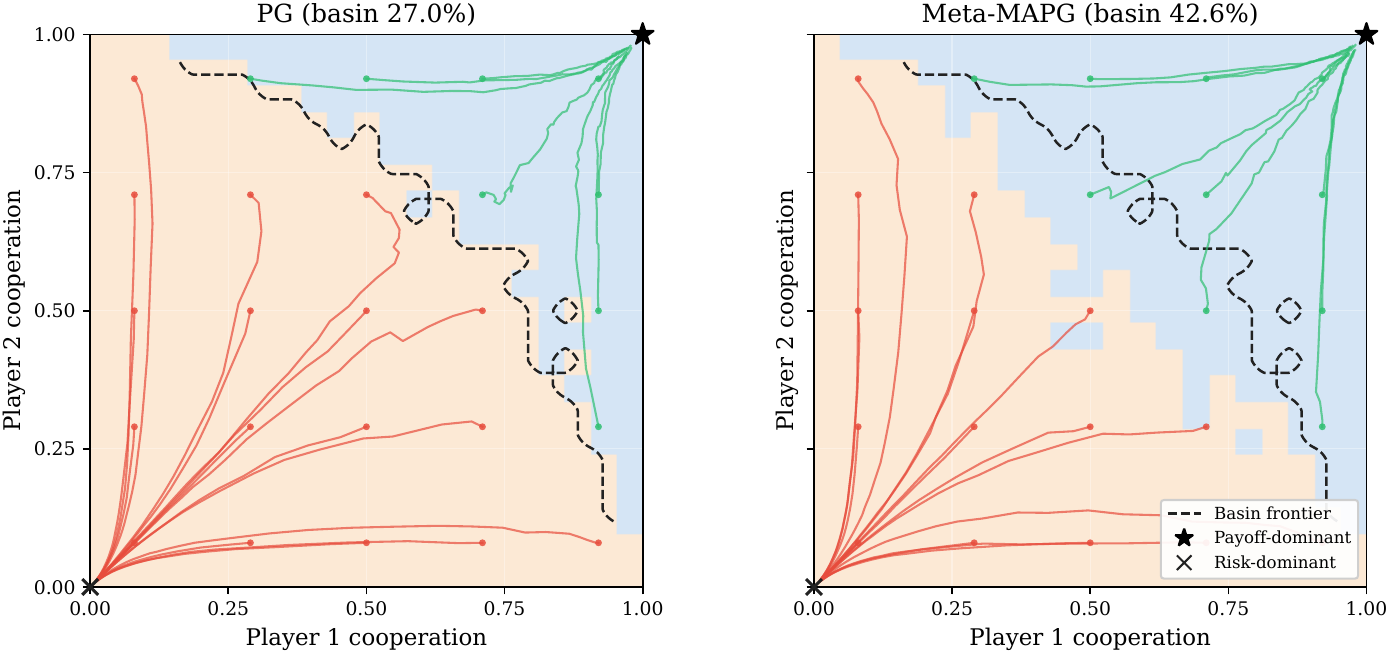}
\caption{Stag Hunt basin geometry under PG (left) and full Meta-MAPG (right). The empirical cooperative basin expands from $27.0\%$ to $42.6\%$ on the $21\times 21$ initialisation grid; the dashed contour is the PG separatrix reused in both panels as a fixed reference, not the Meta-MAPG separatrix.}
\label{fig:basin-geometry}
\end{figure}

Figure~\ref{fig:basin-geometry} is the headline geometric result: on the same $21\times 21$ initialisation grid, the cooperative basin in Stag Hunt grows from $27.0\%$ under ordinary policy gradient to $42.6\%$ under full Meta-MAPG, with the same empirical PG separatrix used as a reference contour in both panels. Trajectories from a $5\times 5$ sub-grid show how individual trajectories respond: initialisations near the separatrix that converge to $(D,D)$ under PG (red) cross into the cooperative basin under Meta-MAPG (green). The rest of this section tests three predictions about how this gain arises: peer learning carries the basin-entry effect; the peer correction is locally aligned with movement toward $(C,C)$; and annealing the correction after entry does not cost coordination.

\subsection{The peer term carries the basin-entry effect}
\label{subsec:peer-effect}

Section~\ref{sec:decomposition} decomposed the Meta-MAPG correction into own-learning and peer-learning terms; Table~\ref{tab:components} reports the four-arm ablation over $100$ paired seeds in both games. Removing the peer term collapses the result to PG: own-only matches PG exactly ($27\%$ in Stag Hunt, $8\%$ in IPD). Removing only the own term preserves the full Meta-MAPG effect: peer-only matches full Meta-MAPG exactly in Stag Hunt ($42\%$ vs $42\%$) and overlaps within Wilson intervals in IPD ($32\%$ vs $37\%$). The operative split is peer-aware versus non-peer in both games. Visually, this is what changes between the two panels of Figure~\ref{fig:basin-geometry}: the gained cells --- those where Meta-MAPG rescues a PG failure --- concentrate along the separatrix, where a directional change in the warm-up update can redirect a trajectory across the boundary.

IPD gives the sharper second test. Without a two-dimensional basin plot, the comparison is Table~\ref{tab:components}: plain PG and own-only stay at $8\%$ cooperative success, while peer-only reaches $32\%$ and full Meta-MAPG reaches $37\%$. The peer-vs-non-peer multiplier is roughly $4$--$5\times$ in IPD, compared with about $1.6\times$ in Stag Hunt. The empirical signal in both games is peer-vs-non-peer rather than full-vs-peer-only; the own-only arm is included for decomposition fidelity, and at the basin-entry stage it tracks PG, consistent with the theory.

\begin{table}[h]
\centering
\small
\begin{tabular}{@{}lcc@{}}
\toprule
Method    & Stag Hunt & IPD \\
\midrule
PG        & $27\%$ $[19.3, 36.4]$ & $\phantom{0}8\%$ $[\phantom{0}4.1, 15.0]$ \\
Own-only  & $27\%$ $[19.3, 36.4]$ & $\phantom{0}8\%$ $[\phantom{0}4.1, 15.0]$ \\
Peer-only & $42\%$ $[32.8, 51.8]$ & $32\%$ $[23.7, 41.7]$ \\
Meta-MAPG & $42\%$ $[32.8, 51.8]$ & $37\%$ $[28.2, 46.8]$ \\
\bottomrule
\end{tabular}
\caption{Component ablation over $100$ paired seeds, with Wilson $95\%$ confidence intervals. Peer-aware methods separate from non-peer methods in both tabular games.}
\label{tab:components}
\end{table}

\subsection{The peer correction is locally aligned with cooperation}
\label{subsec:alignment}

Proposition~\ref{prop:basin} conditions basin enlargement on local alignment of the peer correction with movement toward the target, formally $\mu_M > 0$. We do not estimate the Jacobian or measure $\mu_M$ directly. Instead, the Stag Hunt grid lets us check a directional implication: the target direction is the unit vector toward $(C,C)$, and the first-update peer correction in probability space is $\Delta p_i = \Delta\theta_i\,p_i(1-p_i)$. A cell is \emph{gained} if Meta-MAPG succeeds on at least $50\%$ of seeds and PG on fewer than $50\%$; it is \emph{lost} under the converse. The mask contains $72$ gained cells and no lost cells.

\begin{figure}[t]
\centering
\includegraphics[width=0.7\linewidth]{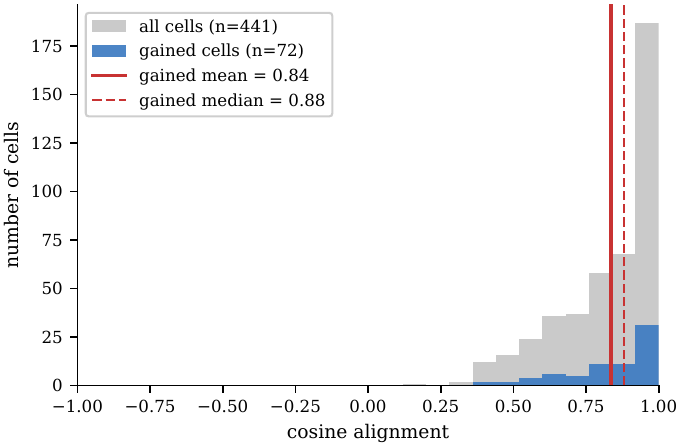}
\caption{Cosine alignment of the first-update peer correction with the direction to $(C,C)$ on the Stag Hunt grid. Grey: all $441$ cells (mean $0.83$). Blue: $72$ gained cells where Meta-MAPG rescues PG (mean $0.84$, median $0.88$, marked). The mass concentrates above $0.7$, with no support below zero, consistent with the directional implication of $\mu_M > 0$ on the cells that drive the basin-entry gain.}
\label{fig:alignment}
\end{figure}

Figure~\ref{fig:alignment} reports cosine alignment of the peer correction with the $(C,C)$ direction. Across the full grid the mean is $0.83$ (median $0.88$); restricted to gained cells the mean is $0.84$ (median $0.88$). Values near $1$ mean the correction points almost exactly toward cooperation. A sizeable mass below zero, or any lost cells, would have contradicted the basin-entry reading; the data show neither. IPD does not have a matching two-dimensional diagnostic here because the policy parameterisation is finite-horizon trajectory-level rather than a Bernoulli unit square, so its role in this section remains the sharper ablation in Section~\ref{subsec:peer-effect}.

\subsection{Cooldown does not cost coordination}
\label{subsec:cooldown}

Constant peer correction shifts the zero-update point of the dynamics by $O(\lambda)$ (Proposition~\ref{prop:basin}(i)), so it need not converge to a Nash of the original game. Proposition~\ref{thm:cooldown} prescribes annealing after basin entry. Over $2000$ Stag Hunt steps with $80$ paired seeds, ordinary PG reaches $31.25\%$ cooperative success, constant Meta-MAPG reaches $41.25\%$, and shape-then-cool reaches $42.50\%$ (Table~\ref{tab:cooldown}). The two Meta-MAPG schedules overlap within Wilson intervals, and second-half cooperation standard deviations stay below $5\times 10^{-4}$ for both. We do not claim cooldown is empirically necessary at this horizon; its role is Nash-target insurance rather than an empirically separated effect. The cooldown experiment is Stag Hunt only because the visible-horizon retention check needs the same Bernoulli parameterisation used elsewhere in this section. The IPD ablation in Section~\ref{subsec:peer-effect} already shows that constant peer correction does not destabilise cooperation in the second game.

\begin{table}[h]
\centering
\small
\begin{tabular}{@{}lccc@{}}
\toprule
Schedule              & Seeds & Cooperative success & Wilson $95\%$ CI \\
\midrule
PG (no shaping)       & $80$ & $31.25\%$ & $[22.2, 42.1]$ \\
Constant Meta-MAPG    & $80$ & $41.25\%$ & $[31.1, 52.2]$ \\
Shape-then-cool       & $80$ & $42.50\%$ & $[32.3, 53.4]$ \\
\bottomrule
\end{tabular}
\caption{Stag Hunt cooldown ablation over $80$ paired seeds. Constant Meta-MAPG and shape-then-cool overlap within Wilson intervals; the gain is retained after annealing. Cooldown is Nash-target insurance under Proposition~\ref{thm:cooldown}, not an empirically separated effect at this horizon.}
\label{tab:cooldown}
\end{table}

\subsection{Robustness and scope}
\label{subsec:tabular-scope}

Three Stag Hunt controls leave the directional reading unchanged. Sweeping $\lambda$ alone raises basin coverage from $29.8\%$ at $\lambda = 0$ to $62.8\%$ at $\lambda = 5$; the effect is monotone in shaping strength rather than a knob choice. Shifting the success threshold across $\tau \in \{0.78, 0.80, 0.82, 0.84, 0.86\}$ preserves the peer-vs-non-peer separation at every cut. A norm-matched PG control, with the PG learning rate scaled to match Meta-MAPG's average early-update norm, leaves PG at $27.2\%$ and Meta-MAPG at $43.3\%$, ruling out an update-magnitude artefact.

The conclusion is narrow. In tabular Stag Hunt and IPD, peer-aware methods separate from non-peer methods on cooperative basin entry, while peer-only and full Meta-MAPG remain indistinguishable within Wilson intervals. The Stag Hunt alignment diagnostic supports the alignment-conditioned basin-entry view as an empirical proxy rather than as a measurement of $\mu_M$. Shape-then-cool is best interpreted as a Nash-preserving handoff. These experiments do not prove Proposition~\ref{prop:basin}, establish scalability, validate a benchmark result, or show that Meta-MAPG always helps. The neural-policy probe in the next section asks whether any of this transfers.

\section{Neural-Policy Probe}
\label{sec:neural-experiments}

We include a small neural-policy probe only to ask whether the peer-versus-non-peer pattern leaves any trace beyond tabular games. All arms share the same IPPO backbone, architecture, optimiser settings, and budget: \texttt{ippo}, \texttt{own\_only}, \texttt{peer\_only}, full \texttt{meta\_mapg}, and fixed-time \texttt{handoff}. Each arm runs $25$ paired seeds on Overcooked \texttt{forced\_coordination}, three MPE tasks, and a PettingZoo IPD fallback; implementation details and full tables are in Appendices~\ref{app:neural-impl}--\ref{app:neural-sweep}.

The honest summary is that one environment carries the entire signal, and that signal is fragile. On Overcooked \texttt{forced\_coordination}, \texttt{peer\_only} has the highest mean final return ($4.77$), above \texttt{ippo} ($3.10$) and full \texttt{meta\_mapg} ($2.71$), but every arm has median zero and the \texttt{peer\_only} mean is driven by one seed reaching $117$. The arms differ in opposite directions: \texttt{ippo} finds the basin more reliably, \texttt{peer\_only} produced the deepest tail learner. We read this as a tail-driven signal consistent with peer-aware corrections occasionally surfacing a deeper coordination basin, not as evidence that Meta-MAPG outperforms IPPO. The \texttt{handoff} failure ($0/25$ learners) does not refute Proposition~\ref{thm:cooldown}, which is conditional on basin entry by handoff time; a fixed-time neural warm-up did not reliably leave the policy in a recoverable basin. The remaining environments do not differentiate: MPE mostly saturates or overlaps across arms, and the IPD fallback collapses to defection.

\section{Conclusion and Limitations}

The standard framing of multi-agent policy-gradient convergence asks whether learning reaches a Nash equilibrium. This paper argues that the prior question --- which Nash equilibrium --- is both well-posed and tractable. By reinterpreting Meta-MAPG as a basin-entry mechanism rather than a rapid-adaptation algorithm, we show that the peer-learning correction is the load-bearing component for equilibrium selection: it is the only part of the update that remains active at Nash and can therefore reshape the geometry of the attraction regions. Under a local alignment condition, this reshaping increases the probability of entering a criterion-selected target basin, and a shape-then-cool schedule recovers convergence to a Nash equilibrium of the original game after entry. The tabular experiments make this visible: in Stag Hunt the cooperative basin grows from $27\%$ to $42\%$, the gain is carried entirely by the peer term, and the alignment diagnostic confirms the geometric reading.

The guarantees are local and conditional. Proposition~\ref{prop:basin} requires $\mu_M > 0$, which holds when the peer correction happens to point toward the target equilibrium --- a condition we can check empirically in tabular settings but cannot enforce in general. Proposition~\ref{thm:cooldown} requires the iterate to have entered the certified attraction region by the handoff time; if warm-up fails to deliver this, cooldown provides no guarantee, as the neural handoff results illustrate. The alignment condition depends on both the game and the initialisation: there is no reason to expect it holds uniformly across the policy space, and the gained-cell analysis in Section~\ref{subsec:alignment} is a local empirical proxy, not a global certificate.

Beyond these formal limitations, three substantive questions remain open. First, criterion choice: the paper takes $W$ as given, but payoff dominance, welfare, fairness, and robustness can conflict, and choosing among them is a normative problem the framework does not address. Second, scaling: the tabular results are clean because the basin geometry is visible; in neural settings the geometry is hidden, the signal is fragile and tail-driven, and it is unclear whether the peer-correction mechanism survives at the scale where MARL is practically relevant. Third, the constructive role of own-learning: we show it is inert at Nash and does not drive basin entry at the warm-up stage, but in long-horizon or non-stationary settings it may contribute along the learning path in ways the current analysis does not capture.

\bibliographystyle{plainnat}
\bibliography{reference}


\appendix

\section{Proof of Proposition~\ref{prop:sa}: Meta-MAPG as Stochastic Approximation}
\label{app:sa}

We prove the statement for the joint stacked update. The proof is blockwise in the agents, and stacking the blocks gives the displayed vector decomposition.

At iteration $n$, let $L_n$ denote the finite unroll length used by the Meta-MAPG estimator. If a fixed unroll length is used, this proof applies with $L_n \equiv L$. Let
\[
\mathcal{F}_n := \sigma(\phi_0, Z_0, \ldots, Z_{n-1})
\]
be the natural filtration generated by the initial parameter and all trajectory bundles used before iteration $n$, where $Z_k$ denotes all sampling randomness used to construct the update at iteration $k$. Thus $\phi_n$ is $\mathcal{F}_n$-measurable, and the new trajectory bundle $Z_n$ is sampled conditionally on $\mathcal{F}_n$ under the policies generated from $\phi_n$.

For agent $i$, write the sampled update estimator in the form
\[
g_{i,n} = \hat{v}_{i,n} + \lambda_n\bigl(\hat{m}^{\mathrm{own}}_{i,L_n,n} + \hat{m}^{\mathrm{peer}}_{i,L_n,n}\bigr),
\]
where $\hat{v}_{i,n}$ is the sampled ordinary policy-gradient part and $\hat{m}^{\mathrm{own}}_{i,L_n,n}$, $\hat{m}^{\mathrm{peer}}_{i,L_n,n}$ are the sampled finite-unroll own-learning and peer-learning terms from the Meta-MAPG decomposition. Define the stacked vectors
\[
g_n := (g_{i,n})_{i=1}^N, \quad v(\phi_n) := \bigl(\nabla_{\phi^i} V_{i,\rho}(\phi_n)\bigr)_{i=1}^N,
\]
and
\[
M_{L_n}(\phi_n) := M^{\mathrm{own}}_{L_n}(\phi_n) + M^{\mathrm{peer}}_{L_n}(\phi_n),
\]
where $M^{\mathrm{own}}_{L_n}$ and $M^{\mathrm{peer}}_{L_n}$ denote the deterministic finite-unroll corrections obtained by taking conditional expectations of the corresponding sampled own-learning and peer-learning estimators.

\subsection*{Step 1: Conditional mean of the sampled estimator}

Let $\bar{g}_n := \mathbb{E}[g_n \mid \mathcal{F}_n]$ be the conditional mean of the sampled update. For the ordinary PG component, the likelihood-ratio identity gives, for each agent $i$,
\[
\mathbb{E}[\hat{v}_{i,n} \mid \mathcal{F}_n] = \nabla_{\phi^i} V_{i,\rho}(\phi_n).
\]
The interchange of gradient and expectation is justified by (RC1)--(RC2): rewards are bounded, policies are twice differentiable, and the relevant score functions and Hessians are uniformly bounded on the local compact set $K$ containing the iterates under consideration.

For the finite-unroll Meta-MAPG correction, applying the same likelihood-ratio identity recursively through the $L_n$ inner-loop updates and using the chain rule gives
\[
\mathbb{E}\bigl[\hat{m}^{\mathrm{own}}_{i,L_n,n} + \hat{m}^{\mathrm{peer}}_{i,L_n,n} \mid \mathcal{F}_n\bigr] = M^{\mathrm{own}}_{i,L_n}(\phi_n) + M^{\mathrm{peer}}_{i,L_n}(\phi_n) + \delta_{i,n},
\]
where $\delta_{i,n}$ is the deterministic finite-unroll truncation residual. For an exact finite-unroll target, $\delta_{i,n} = 0$. When the finite unroll is used as an approximation to the untruncated meta-gradient, (RC4) gives a deterministic sequence $\beta_n \to 0$ and a constant $C < \infty$ such that, after stacking over agents,
\[
\|\delta_n\| := \bigl\|(\delta_{i,n})_{i=1}^N\bigr\| \leq C\beta_n.
\]
The stronger form used later in Proposition~\ref{thm:cooldown} is $\sum_n \alpha_n \beta_n < \infty$. Stacking the two identities and absorbing the deterministic residual $\lambda_n \delta_n$ into $b_n$, we obtain
\[
\bar{g}_n = v(\phi_n) + \lambda_n M_{L_n}(\phi_n) + b_n,
\]
with $\|b_n\| \leq C\beta_n$.

\subsection*{Step 2: Martingale-difference noise}

Define $\xi_{n+1} := g_n - \mathbb{E}[g_n \mid \mathcal{F}_n] = g_n - \bar{g}_n$. Then, by construction, $\mathbb{E}[\xi_{n+1} \mid \mathcal{F}_n] = 0$, so $\{\xi_{n+1}\}$ is a martingale-difference sequence with respect to $\{\mathcal{F}_{n+1}\}$.

\subsection*{Step 3: Conditional second-moment bound}

By (RC3), used in its uniform-on-$K$ form for the sampled finite-unroll likelihood-ratio estimator, there exists $\sigma^2 < \infty$ such that
\[
\mathbb{E}\bigl[\|g_n - \mathbb{E}[g_n \mid \mathcal{F}_n]\|^2 \mid \mathcal{F}_n\bigr] \leq \sigma^2 \quad \text{for all } n.
\]
The uniformity is essential when $L_n$ grows; it cannot in general be derived from the discount factor $\gamma_{\mathrm{disc}}$ alone, because $\gamma_{\mathrm{disc}}$ discounts environment time, not the length of the meta-learning unroll.

\subsection*{Step 4: Stochastic-approximation decomposition}

Combining the conditional mean and noise definitions, we obtain
\[
g_n = v(\phi_n) + \lambda_n M_{L_n}(\phi_n) + \xi_{n+1} + b_n,
\]
where $\mathbb{E}[\xi_{n+1} \mid \mathcal{F}_n] = 0$, $\mathbb{E}[\|\xi_{n+1}\|^2 \mid \mathcal{F}_n] \leq \sigma^2$, $\|b_n\| \leq C\beta_n$, and $\beta_n \to 0$. This proves Proposition~\ref{prop:sa}.
\hfill$\square$


\section{Proof of Proposition~\ref{prop:basin}: Basin Geometry Under Peer Learning}
\label{app:basin}

We prove the three claims for the shaped vector field
\[
    F_\lambda(\phi) := v(\phi)+\lambda M_L(\phi),
\]
where $M_L$ is the finite-unroll Meta-MAPG correction. If the statement is specialised to the peer-learning correction only, the same proof applies with $M_L$ replaced everywhere by $M_L^{\mathrm{peer}}$.

\subsection*{Step 1: Consequences of the SOS condition}

Let $A := Dv(\phi^*)$, $B := DM_L(\phi^*)$, $S_A := (A+A^\top)/2$, $S_B := (B+B^\top)/2$. Since $\phi^*$ is an SOS Nash equilibrium of $v$ with constants $\mu > 0$ and $r > 0$, we have $S_A \preceq -\mu I$ and $A$ is nonsingular. The peer-alignment quantity $\mu_M > 0$ is equivalent to $S_B \preceq -\mu_M I$.

\subsection*{Step 2: Fixed-point shift}

By the implicit function theorem applied to $H(\phi,\lambda) := v(\phi)+\lambda M_L(\phi)$ at $(\phi^*,0)$, there exists a unique $C^1$ curve $\lambda \mapsto \phi^*_\lambda$ with $F_\lambda(\phi^*_\lambda)=0$ and $\phi^*_0=\phi^*$. Differentiating at $\lambda=0$ gives $\left.\frac{d\phi^*_\lambda}{d\lambda}\right|_{\lambda=0} = -A^{-1}M_L(\phi^*)$, from which $\|\phi^*_\lambda-\phi^*\| = O(\lambda)$. This proves part~(i).

\subsection*{Step 3: Local drift improvement}

Combining $S_A \preceq -\mu I$ and $S_B \preceq -\mu_M I$, the symmetrised Jacobian of $F_\lambda$ at $\phi^*_\lambda$ satisfies $(DF_\lambda(\phi^*_\lambda)+DF_\lambda(\phi^*_\lambda)^\top)/2 \preceq -\tfrac{3}{4}(\mu+\lambda\mu_M)I$ for small $\lambda$. Via the fundamental theorem of calculus and a Lipschitz bound on $DF_\lambda$, for all $\phi \in B_{\rho_\lambda}(\phi^*_\lambda)$ with $\rho_\lambda := (\mu+\lambda\mu_M)/(2L_\lambda)$,
\[
\langle F_\lambda(\phi),\phi-\phi^*_\lambda\rangle \leq -\tfrac{1}{2}(\mu+\lambda\mu_M)\|\phi-\phi^*_\lambda\|^2.
\]
This proves part~(ii).

\subsection*{Step 4: Certified basin expansion}

The inequality above is an SOS drift certificate for $F_\lambda$ around $\phi^*_\lambda$ with radius $\rho_\lambda$. Using a common local Lipschitz constant $L_{\mathrm{loc}}$ for both fields, the ordinary PG certified radius is $\rho_0 = \mu/(2L_{\mathrm{loc}})$ while the shaped field has $\widetilde{\rho}_\lambda = (\mu+\lambda\mu_M)/(2L_{\mathrm{loc}}) > \rho_0$ when $\mu_M > 0$ and $\lambda > 0$. This proves part~(iii), completing the proof of Proposition~\ref{prop:basin}.
\hfill$\square$


\section{Proof of Proposition~\ref{thm:cooldown}: Local Convergence After Cooldown}
\label{app:cooldown}

We prove that, once the post-handoff iterate has entered the certified attraction region of the original SOS Nash equilibrium, the cooldown correction contributes only a summable perturbation to ordinary PG. Hence the asymptotic dynamics are governed by the original vector field $v$, and the limiting point is $\phi^*$ rather than the shifted zero of $v+\lambda M_L$.

\subsection*{Step 1: Post-handoff dynamics and local bounds}

Let $B := B_{r_{\mathrm{att}}}(\phi^*)$ be the certified attraction region. Work on the successful-entry event $\mathcal{E}_T := \{\phi_T\in B\}$. For $n\geq T$, Proposition~\ref{prop:sa} gives the sampled update with deterministic post-handoff perturbations $q_n := \lambda_n M_{L_n}(\phi_n)+b_n$. Define $e_n := \phi_n-\phi^*$ and $X_n := \|e_n\|^2$.

\subsection*{Step 2: One-step Lyapunov inequality}

Taking conditional expectation and applying the SOS drift condition, Young's inequality for the perturbation term, and local bounds for the second-moment term, gives
\begin{equation}
    \mathbb{E}[X_{n+1}\mid\mathcal{F}_n]
    \leq
    (1-\mu\alpha_n)X_n
    +
    \frac{\alpha_n}{\mu}\|q_n\|^2
    +
    C_1\alpha_n^2.
\end{equation}

\subsection*{Step 3: Summability of the perturbation terms}

From $\|q_n\| \leq B_M\lambda_n+C\beta_n$, together with the cooldown condition $\sum_{n\geq T}\alpha_n\lambda_n<\infty$ and the bias summability $\sum_{n\geq T}\alpha_n\beta_n<\infty$, we get $\sum_{n\geq T}\alpha_n\|q_n\|^2 < \infty$. Combined with $\sum_{n\geq T}\alpha_n^2<\infty$, the remainder terms $r_n$ satisfy $\sum_{n\geq T} r_n < \infty$.

\subsection*{Step 4: Robbins--Siegmund convergence}

Applying the Robbins--Siegmund almost-supermartingale lemma to $(1-\mu\alpha_n)X_n + r_n$, $X_n$ converges almost surely to a finite limit and $\sum_{n\geq T}\alpha_n X_n < \infty$ almost surely. Since $\sum_{n\geq T}\alpha_n=\infty$, the only possible limit is $X_\infty=0$, hence $\phi_n \to \phi^*$ almost surely.
\hfill$\square$


\section{Neural-Policy Implementation Details}
\label{app:neural-impl}

\subsection{IPPO backbone}

All five arms share an Independent PPO (IPPO) backbone. Each agent $i$ maintains a separate actor $\pi_{\theta_i}$ and critic $V_{\phi_i}$. Rollouts of length $L = 256$ (or $400$ for Overcooked) are collected per training iteration. Advantages are computed via Generalised Advantage Estimation with $\gamma_{\text{disc}} = 0.99$ and $\lambda_{\text{GAE}} = 0.95$. The PPO clipped surrogate is used with $\epsilon_{\text{clip}} = 0.2$, entropy coefficient $0.01$, and gradient-norm clip $0.5$. Each rollout uses $4$ PPO epochs with minibatch size $256$. Actor learning rate is $3 \cdot 10^{-4}$; critic learning rate is $1 \cdot 10^{-3}$.

\subsection{Correction terms}

The own- and peer-learning corrections are estimated using the DiCE estimator \citep{foerster2018dice}, which provides a differentiable Monte Carlo surrogate for higher-order policy gradients through a learning step. With inner step $\eta_{\text{inner}} = 0.1$ and advantage estimate $\hat{A}_i$, the per-agent correction terms are
\[
\Delta^{\text{own}}_i = \nabla_{\theta_i} \log \pi_{\theta_i}(a_i \mid s) \cdot \hat{A}_i, \qquad
\Delta^{\text{peer}}_i = \nabla_{\theta_i} \log \pi_{\theta_{-i}}(a_{-i} \mid s) \cdot \hat{A}_i.
\]
Each arm activates a different subset: \texttt{ippo} uses $\Delta_{\text{PG}}$ only; \texttt{own\_only} adds $\Delta^{\text{own}}$; \texttt{peer\_only} adds $\Delta^{\text{peer}}$; \texttt{meta\_mapg} adds both; \texttt{handoff} runs as \texttt{meta\_mapg} until step $T_{\text{warm}}$ and then disables corrections (revert to \texttt{ippo}). All corrections are norm-clipped to prevent blow-up:
\[
\|\Delta_{\text{corr}}\|_2 \leq c \cdot \|\Delta_{\text{PG}}\|_2, \qquad c = 1.0.
\]
This is a PPO-compatible practical proxy for the decomposition in Section~\ref{sec:decomposition}, not a theorem-level object: the truncated correction is a deep-MARL surrogate that preserves the own-vs-peer distinction while remaining numerically tractable at neural scale.

\subsection{Paired-seed protocol}

For each seed value $\in \{0, 1, \ldots, 24\}$, all five arms initialise actor and critic weights from the same PRNG state. Differences in final return at fixed seed are therefore attributable to the algorithmic correction, not initialisation noise. The same protocol is used for all five environments.

\begin{table}[h]
\centering
\small
\begin{tabular}{llccccc}
\toprule
Benchmark & Env id & Arms & Seeds & Steps & $T_{\text{warm}}$ & $\tau$ \\
\midrule
MPE & \texttt{simple\_spread} & 5 & 25 & 750k & 200k & $-25.0$ \\
MPE & \texttt{simple\_reference} & 5 & 25 & 750k & 200k & $-22.0$ \\
MPE & \texttt{simple\_speaker\_listener} & 5 & 25 & 750k & 200k & $-10.0$ \\
Overcooked-AI & \texttt{forced\_coordination} & 5 & 25 & 1M & 500k & $3.0$ \\
Melting Pot$^\dagger$ & \texttt{prisoners\_dilemma\_in\_the\_matrix} & 4 & 25 & 1M & 400k & $0.5$ \\
\bottomrule
\end{tabular}
\caption{Per-environment configuration. $T_{\text{warm}}$ applies only to the \texttt{handoff} arm; $\tau$ is the basin-entry threshold used for learner counts. $^\dagger$The \texttt{own\_only} arm is omitted on this benchmark; the \texttt{dm\_meltingpot} substrate was unavailable in the runtime environment, and we use the PettingZoo Iterated Prisoner's Dilemma fallback as a matrix-style proxy.}
\end{table}

\subsection{Sparse-reward fix in Overcooked}

The Overcooked-AI environment wrapper originally returned only the sparse soup-delivery reward ($+20$), discarding the per-step shaped reward (onion pickup, dish handling). Without shaping, vanilla PPO cannot learn \texttt{forced\_coordination} within $1$M steps. We patched the wrapper to add $0.5 \cdot \text{shaped}_i$ to the per-agent reward (the Overcooked-AI default coefficient), with the sparse soup reward split equally between the two cooks. The patch was applied uniformly across all reported Overcooked results.

\subsection{Pilot-threshold cascade}

The pilot stage ran post-patch and produced thresholds $\tau = 0.0$ for Overcooked and the IPD fallback. The full sweep used these stale thresholds, so success-fraction summaries on those two environments report effectively any-non-zero return as success. The honest signal is in mean final return and per-seed distribution, which are what we report in the main text and in Appendix~\ref{app:neural-sweep}.

\subsection{Evaluation and compute}

Every ${\sim}25$k environment steps we run $50$ evaluation episodes with greedy (\textsc{argmax}) action selection and report mean episode return. A seed counts as a learner if its final-checkpoint mean return crosses the benchmark-specific threshold $\tau$. Greedy evaluation underestimates stochastic-policy performance; this affects all arms equally and is unlikely to change relative orderings. All training was performed on $4\times$ NVIDIA Tesla V100-SXM2-32GB GPUs with $12$ parallel workers via \texttt{multiprocessing.spawn}; total compute is approximately $350$ GPU-hours.

\section{Full Neural-Sweep Tables}
\label{app:neural-sweep}

This appendix provides the complete per-environment final-return distributions that bound the scope of the main-text Overcooked claim. We organise environments from least- to most-differentiating.

\subsection{MPE \texttt{simple\_spread}: saturated}

\begin{table}[h]
\centering
\small
\begin{tabular}{lcccccccc}
\toprule
Arm & $n$ & mean & std & min & p25 & median & p75 & max \\
\midrule
\texttt{ippo}       & 25 & $-22.02$ & $0.75$ & $-23.87$ & $-22.55$ & $-22.08$ & $-21.43$ & $-20.48$ \\
\texttt{own\_only}  & 25 & $-21.85$ & $0.55$ & $-22.80$ & $-22.30$ & $-21.86$ & $-21.35$ & $-20.65$ \\
\texttt{peer\_only} & 25 & $-21.81$ & $0.77$ & $-23.46$ & $-22.18$ & $-21.68$ & $-21.48$ & $-20.01$ \\
\texttt{meta\_mapg} & 25 & $-21.80$ & $0.55$ & $-23.13$ & $-22.21$ & $-21.74$ & $-21.47$ & $-20.67$ \\
\texttt{handoff}    & 25 & $-21.86$ & $0.66$ & $-23.05$ & $-22.12$ & $-21.83$ & $-21.69$ & $-19.75$ \\
\bottomrule
\end{tabular}
\caption{\texttt{simple\_spread}, $\tau = -25$. All arms reach $25/25$ learners; mean returns differ by less than within-arm standard deviation. Median first-hit step is identical across arms.}
\end{table}

The 3-agent / 3-landmark cooperative navigation task is fully saturated. This benchmark serves as a control: opponent-aware corrections do not hurt on easy cooperative tasks, but neither do they help.

\subsection{MPE \texttt{simple\_reference}: weakly suggestive}

\begin{table}[h]
\centering
\small
\begin{tabular}{lcccccccc}
\toprule
Arm & $n$ & mean & std & min & p25 & median & p75 & max \\
\midrule
\texttt{ippo}       & 25 & $-21.98$ & $3.87$ & $-36.20$ & $-22.14$ & $-20.81$ & $-19.74$ & $-18.79$ \\
\texttt{own\_only}  & 25 & $-22.68$ & $4.96$ & $-38.19$ & $-22.60$ & $-21.02$ & $-19.90$ & $-19.24$ \\
\texttt{peer\_only} & 25 & $-22.21$ & $3.36$ & $-31.39$ & $-24.07$ & $-21.06$ & $-19.64$ & $-18.85$ \\
\texttt{meta\_mapg} & 25 & $-22.21$ & $4.21$ & $-37.70$ & $-22.08$ & $-21.16$ & $-19.78$ & $-19.02$ \\
\texttt{handoff}    & 25 & $-22.87$ & $4.94$ & $-42.38$ & $-22.93$ & $-21.26$ & $-20.29$ & $-18.95$ \\
\bottomrule
\end{tabular}
\caption{\texttt{simple\_reference}, $\tau = -22$. Larger seed variance reveals tail behaviour; learner counts span $16$--$18/25$ with overlapping Wilson intervals.}
\end{table}

The 2-agent communication task introduces real seed variance ($\text{std} \approx 4$ vs $0.5$--$0.8$ on \texttt{simple\_spread}). \texttt{peer\_only} exhibits the smallest within-arm dispersion ($\text{std} = 3.36$); \texttt{handoff} shows the worst-case tail seed ($\min = -42.38$). Wilson intervals on learner counts overlap heavily and the nominal ordering does not match the mean-return ordering.

\subsection{MPE \texttt{simple\_speaker\_listener}: equal failure}

\begin{table}[h]
\centering
\small
\begin{tabular}{lcccccccc}
\toprule
Arm & $n$ & mean & std & min & p25 & median & p75 & max \\
\midrule
\texttt{ippo}       & 25 & $-25.81$ & $1.96$ & $-29.87$ & $-27.44$ & $-25.46$ & $-24.22$ & $-22.10$ \\
\texttt{own\_only}  & 25 & $-25.82$ & $2.05$ & $-30.73$ & $-27.71$ & $-25.21$ & $-24.37$ & $-21.82$ \\
\texttt{peer\_only} & 25 & $-25.77$ & $1.95$ & $-29.84$ & $-27.15$ & $-25.65$ & $-24.36$ & $-21.58$ \\
\texttt{meta\_mapg} & 25 & $-25.88$ & $2.04$ & $-29.72$ & $-27.45$ & $-25.83$ & $-24.26$ & $-21.71$ \\
\texttt{handoff}    & 25 & $-25.64$ & $2.01$ & $-28.97$ & $-27.71$ & $-25.54$ & $-24.13$ & $-21.81$ \\
\bottomrule
\end{tabular}
\caption{\texttt{simple\_speaker\_listener}, $\tau = -10$. All arms fail to cross threshold within $750$k env-steps; final returns are near-identical across arms.}
\end{table}

The asymmetric speaker-listener communication task is uniformly failed across all five arms. All arms produce near-identical statistics; differences are about one seed wide.

\subsection{Overcooked \texttt{forced\_coordination}: super-bimodal}

\begin{table}[h]
\centering
\small
\begin{tabular}{lcccccccc}
\toprule
Arm & $n$ & mean & std & min & p25 & median & p75 & max \\
\midrule
\texttt{ippo}       & 25 & $3.10$ & $\phantom{0}9.23$ & $0.00$ & $0.00$ & $0.00$ & $0.00$ & $\phantom{0}34.50$ \\
\texttt{own\_only}  & 25 & $0.06$ & $\phantom{0}0.20$ & $0.00$ & $0.00$ & $0.00$ & $0.00$ & $\phantom{00}0.75$ \\
\texttt{peer\_only} & 25 & $4.77$ & $22.91$ & $0.00$ & $0.00$ & $0.00$ & $0.00$ & $117.00$ \\
\texttt{meta\_mapg} & 25 & $2.71$ & $\phantom{0}8.76$ & $0.00$ & $0.00$ & $0.00$ & $0.00$ & $\phantom{0}33.00$ \\
\texttt{handoff}    & 25 & $0.21$ & $\phantom{0}0.58$ & $0.00$ & $0.00$ & $0.00$ & $0.00$ & $\phantom{00}2.25$ \\
\bottomrule
\end{tabular}
\caption{\texttt{forced\_coordination}, $\tau = 3.0$. Median final return is $0.00$ for every arm; distributions are super-bimodal (most seeds fail entirely, rare learners reach high returns). The mean is dragged by tail learners.}
\end{table}

\texttt{forced\_coordination} is the strongest coordination test in the suite: the two cooks are physically separated by an interior counter, soup cannot be made by either alone, and partial credit is awarded only via shaped reward (onion pickup, dish handling, soup delivery). Coordinated soup delivery requires the agents to converge on a sequential convention. The full per-arm distribution is asymmetric across arms in two respects. \texttt{peer\_only} produced one seed reaching $117$ --- a fully-synchronised partnership delivering multiple soups per evaluation episode --- with all $24$ remaining seeds at zero or near-zero. \texttt{ippo} produced four learners between $3$ and $34.5$, no deeper than that. \texttt{own\_only} produced no learners and a maximum of $0.75$ across all $25$ seeds; \texttt{handoff} produced no learners and a maximum of $2.25$. These distributions motivate the main-text reading of ``fragile, tail-driven differentiating signal''.

\subsection{PettingZoo IPD fallback: equilibrium collapse}

\begin{table}[h]
\centering
\small
\begin{tabular}{lcccccccc}
\toprule
Arm & $n$ & mean & std & min & p25 & median & p75 & max \\
\midrule
\texttt{ippo}       & 25 & $0.00$ & $0.00$ & $0.00$ & $0.00$ & $0.00$ & $0.00$ & $0.00$ \\
\texttt{peer\_only} & 25 & $0.00$ & $0.00$ & $0.00$ & $0.00$ & $0.00$ & $0.00$ & $0.00$ \\
\texttt{meta\_mapg} & 25 & $0.00$ & $0.00$ & $0.00$ & $0.00$ & $0.00$ & $0.00$ & $0.00$ \\
\texttt{handoff}    & 25 & $0.00$ & $0.00$ & $0.00$ & $0.00$ & $0.00$ & $0.00$ & $0.00$ \\
\bottomrule
\end{tabular}
\caption{\texttt{prisoners\_dilemma\_in\_the\_matrix\_\_repeated} (PettingZoo IPD fallback), $\tau = 0.5$. All arms collapse to mutual defection; mean and std are zero across arms.}
\end{table}

The PettingZoo IPD fallback equilibrium-collapses to mutual defection, the Nash equilibrium for short-horizon IPD without communication or memory. Median first-hit on this benchmark is the very first eval checkpoint (step $512$); seeds transiently exhibit cooperative spikes in the early random-policy regime, then descend to defection. This benchmark is uninformative for our hypothesis. The intended \texttt{dm\_meltingpot} \texttt{prisoners\_dilemma\_in\_the\_matrix} substrate is a 2D-grid spatial coordination task that would have provided a non-trivial signal; it was unavailable in our runtime.

\end{document}